\documentclass[runningheads]{llncs}
\usepackage[T1]{fontenc}
\usepackage{graphicx}
\usepackage{booktabs}

\usepackage[textsize=small,colorinlistoftodos]{todonotes}

\usepackage[hyphens]{url} 
\usepackage{graphicx} 
\usepackage{graphicx}  

\usepackage{caption}  
\usepackage{bm}
\usepackage{amssymb,amsmath,bbold}
\usepackage{subcaption}
\usepackage{minted}
\usepackage{stfloats}  
\usepackage{booktabs}

\usepackage{array}

\usepackage[hidelinks]{hyperref} 
\usepackage{varioref}
\usepackage{todonotes}

\usepackage{mwe}

\begin{document}

\title{Counterfactual Learning of New Adaptive Instructional Policies using Logged Data}

\toctitle{Counterfactual Learning of New Adaptive Instructional Policies using Logged Data}

\titlerunning{Counterfactual Learning of Adaptive Policies using Logged Data}

\author{Samuel Girard\inst{1,2} \and
Sein Minn\inst{3} \and
Amel Bouzeghoub\inst{4} \and
Jill-Jênn Vie\inst{1}}

\authorrunning{Girard et al.}

\tocauthor{Samuel~Girard, Sein~Minn, Amel~Bouzeghoub, Jill-Jênn~Vie}

\institute{Soda Team, Inria Saclay, Palaiseau, France \email{\{samuel.girard,jill-jenn.vie\}@inria.fr}
\and
Pix, Paris, France
\and
Asian Institute of Technology, Khlong Nueng, Thailand 
\email{sein-minn@ait.asia}
\and
SAMOVAR, Télécom SudParis, Institut Polytechnique de Paris, Palaiseau, France
\email{amel.bouzeghoub@telecom-sudparis.eu}}

\maketitle

\begin{abstract}
Optimizing instructional policies in Intelligent Tutoring Systems (ITS) typically requires costly online experimentation or student simulators that may fail to capture real-world dynamics. This paper introduces an offline contextual bandit framework that learns new adaptive policies directly from logged interaction data. By mapping student-item interactions onto a continuous latent proficiency-difficulty scale using a Rasch model, we cast the tutoring process as a continuous stochastic bandit problem. We propose a novel reward function designed to optimize ``flow'' by balancing task challenge with student success. Our approach includes a round-specific behavior policy estimation that serves as both a propensity model for off-policy evaluation and a diagnostic tool for ITS adaptivity. We demonstrate the efficacy of this framework across four large-scale real-world datasets, achieving consistent policy improvements over the logged behavior policy. The results show that effective instructional policies can be learned and visualized within seconds of computation, providing a scalable path for improving adaptive learning systems without further data collection.
\end{abstract}

\section{Introduction}

Adaptive learning aims to present each student with exercises that match their current ability. In intelligent tutoring systems (ITS), this means assigning tasks that are neither too easy, which can lead to disengagement, nor too difficult, which can cause failure and frustration~\cite{bjork2011making,The_zones_of_proximal_flow}. Calibrating instructional difficulty is therefore central to maintaining engagement and supporting learning~\cite{Clement2015,fang2019meta}.
Counterfactual learning is a machine learning approach that uses hypothetical ``what-if'' scenarios (called \emph{counterfactuals}) to estimate the effect of interventions or actions that were not actually observed.
This method is especially attractive in settings where online experimentations on students may be costly, or raise ethical issues~\cite{levine2020offlinereinforcementlearningtutorial,gao2023hopehumancentricoffpolicyevaluation,gao2024on,yang2025themes}.
Because ITS collect large logs of student--exercise interactions, they offer the possibility of improving tutoring policies directly from data. Still, policies optimized in simulation may overfit the student model used to generate synthetic trajectories~\cite{doroudi2017robust}, so in this paper we are interested in model-free techniques such as inverse propensity scoring as well. 

In this paper, we study offline learning and evaluation of difficulty-assignment policies from logged tutoring data. Rather than modeling the full tutoring process, we adopt a reduced-form view in which each interaction is represented on a continuous latent proficiency--difficulty scale. Conceptually, the quantity of interest is a stochastic policy that maps latent student proficiency to a distribution over exercise difficulty. Since true proficiency is unobserved, we do not estimate this object directly; instead, we infer a one-dimensional Rasch proficiency proxy from the log and use it to approximate the corresponding reduced-form policy over difficulty~\cite{rasch1960studies}. This reduction yields an offline contextual bandit problem with continuous context and continuous action. It also allows us to estimate the reduced-form logging policy from the data. Importantly, this estimated policy is not the ITS's internal decision rule; rather, it captures the effective allocation pattern observed in the log. As such, it is useful not only for propensity estimation in off-policy evaluation, but also as a practical diagnostic of whether the ITS behaves adaptively across students and rounds.
We define a reward that favors successful responses to more difficult items, thereby operationalizing the idea of desirable difficulty and its connection to flow~\cite{csikszentmihalyi1992optimal,bjork2011making,The_zones_of_proximal_flow}. The reward is computed directly from logged outcomes and item difficulties, so it provides an observable proxy for a pedagogically desirable balance between challenge and attainability, rather than a direct measure of long-term learning gains.



Our framework learns from tutoring logs through a continuous
proficiency--difficulty representation, a reward based on flow computed
from observed responses, and round-specific reduced-form logging
policies for off-policy evaluation. This enables policy evaluation and
learning directly from historical data, without interaction with
students or reliance on a student simulator~\cite{doroudi2017robust,levine2020offlinereinforcementlearningtutorial}.
Because both the logging policy and the learned policies are defined on
a continuous latent scale, they can also be visualized directly, which
is useful for practitioners seeking both improved policies and a
clearer picture of how difficulty is allocated in practice.

Our three contributions are: (i) a reward inspired by flow and
desirable difficulty that incentivizes assigning the hardest exercise a
student can still solve; (ii) a reduced-form behavior-policy estimation
framework, based on continuous context and continuous actions,
that reveals how difficulty allocation varies with student
proficiency and evolves across rounds, providing an interpretable
diagnostic of ITS adaptivity; and (iii) an offline policy learning
framework that combines this behavior model with several off-policy
estimators to learn and compare instructional policies from purely
logged data. 

\section{Related Work}
\label{related-work}

\paragraph{Knowledge tracing and student modeling.}
A large body of work in educational data mining focuses on modeling student knowledge from interaction logs, including knowledge tracing~\cite{corbett1994knowledge}, Bayesian Knowledge Tracing, Item Response Theory, and related latent-variable approaches. These models are typically used to predict future performance or mastery, rather than to optimize exercise assignment directly. Our work uses this literature differently: instead of building a deep sequential student model, we use a simple Rasch representation to place students and items on a common latent proficiency--difficulty scale, which serves as a reduced-form basis for offline policy evaluation and learning.

\paragraph{Counterfactual learning in education.}
Counterfactual evaluation is much less common in education than predictive modeling, because it requires reasoning about outcomes under interventions not observed in the data. Zhao and Heffernan study binary educational interventions, such as the effect of hints on course completion~\cite{zhao2017estimating}. Doroudi et al.\ propose a robust framework for evaluating instructional policies using student models~\cite{doroudi2017robust}. Our setting differs from prior educational work in two ways: we learn directly from logged tutoring data without relying on a student simulator, and we treat exercise assignment as a continuous difficulty-selection problem rather than as a small discrete intervention space.

\paragraph{Model-based versus model-free offline learning.}
Most of student modeling in education is model-based: policies are optimized using a learned model of student evolution, such as HOT-DINA, BKT, multidimensional IRT, or other partially observable Markov decision processes~\cite{subramanian2021deep,yessad2022personalizing,lan2016contextual,Whitehill2017}. While such approaches can capture richer dynamics, they depend heavily on the fidelity of the student model, and policies learned in simulation may fail to transfer to real students~\cite{doroudi2017robust}. Our approach allows model-free learning. We do not attempt to model the full tutoring dynamics; instead, we formulate logged exercise assignment as an offline contextual bandit problem on a latent proficiency--difficulty scale, which enables direct off-policy evaluation and learning from observed data.

\paragraph{Behavior policy estimation and reward design.}
In most off-policy evaluation work, the logging policy is treated mainly as a technical object needed for importance weighting~\cite{levine2020offlinereinforcementlearningtutorial}. In our setting, estimating the reduced-form logging policy also has independent practical value: it reveals how assigned difficulty varies with proficiency and across rounds, providing a diagnostic of the adaptivity expressed in the logged data. On the reward side, prior work has often relied on metrics based on the student model, or sparse long-term objectives such as post-test scores~\cite{subramanian2021deep,Bassen2020,Vassoyan2023}. We instead focus on an immediate reward aligned with desirable difficulty~\cite{bjork2011making} and flow~\cite{csikszentmihalyi1992optimal}, which is directly computable from logged responses and therefore compatible with model-free off-policy evaluation.

\paragraph{Estimators for off-policy evaluation.}
Our estimators build on the off-policy evaluation (OPE) literature for
contextual bandits, which estimates the value a target policy \emph{would} have
obtained had it been deployed, using only data logged under a different
behavior policy~\cite{bottou2013counterfactualreasoninglearningsystems}. Inverse propensity scoring
(IPS) is unbiased but high-variance; its self-normalized variant (SNIPS)
trades a small bias for lower variance and underpins counterfactual risk
minimization from logged feedback~\cite{swaminathan15}.
The doubly robust (DR) estimator combines a reward model with importance
weighting and stays consistent if either is correct~\cite{dudik2011doublyrobustpolicyevaluation},
while mixture importance sampling stabilizes weights across pooled logging
policies via a shared denominator~\cite{Agarwal_2017}.
We adapt these estimators to a continuous proficiency--difficulty action space
with round-specific logging policies, and use their agreement as a reliability
diagnostic.

\section{Method}
\label{method}

Our framework operates in three steps: (i)~project the tutoring log onto a continuous proficiency--difficulty scale, (ii)~define a flow-aligned reward on that scale, and (iii)~estimate a round-specific behavior policy that will serve as the propensity model for offline evaluation and learning.

\subsection{Continuous latent representation}

We assume that each student $u$ has a fixed scalar proficiency
$z_u\in\mathbb{R}$ and each item $i$ has a scalar difficulty
$d_i\in\mathbb{R}$. Correctness is governed by a \emph{response function}
$q:\mathbb{R}^2\to[0,1]$ that maps a proficiency--difficulty pair to the
probability of a correct answer, so that the observed binary correctness
$Y\in\{0,1\}$ satisfies $\Pr(Y{=}1\mid z_u,d_i)=q(z_u,d_i)$. Rather than
defining the action as a discrete item identity, we lift it to the continuous
difficulty level $a_{u,t}:=d_{i_{u,t}}\in\mathcal{A}\subset\mathbb{R}$. This
continuous formulation is better suited to large item banks: it avoids a
high-dimensional discrete action space and enables policies to generalize
across different exercises with similar difficulty. This task can
therefore be seen as a continuous contextual bandit problem.

We estimate proficiencies via the 1PL Rasch model, $q(z,d)=\sigma(z-d)$, where
$\sigma$ is the logistic function and each item is described by a single
difficulty parameter, so that the probability of a correct response depends
only on the gap between ability and difficulty. Fitting this model by maximum
likelihood yields estimates $\hat{z}_u$. The context is then
$x_{u,t}:=\hat{z}_u$ for all $t$; a fixed proxy for latent proficiency rather
than a dynamic knowledge state. Each student trace is thus recast as a sequence
$\bigl(x_{u,t},\,a_{u,t},\,y_{u,t}\bigr)_{t=1}^{T_u}$ on a latent scale, where
$y_{u,t}\in\{0,1\}$ records whether student $u$ answered correctly at round $t$.
This introduces two error sources: (i)~a \emph{representation error}, since true
correctness may not reduce to a single scalar pair; and (ii)~a \emph{model
bias}, since the Rasch form $\sigma(z-d)$ may not match the true $q$.

This reduction is well suited to our objective defined below: the flow-aligned reward is
immediate by construction, so we do not need to perform long-horizon credit
assignment, and a model-free bandit formulation avoids the dependence
on a student simulator that can cause simulation-trained policies to
fail on real students~\cite{doroudi2017robust}.

\subsection{Reward design}

Our reward operationalizes \emph{desirable difficulty}~\cite{bjork2011making}: exercises should be challenging but attainable, steering students toward a state of \emph{flow}~\cite{csikszentmihalyi1992optimal}. Given an observed outcome $y_{u,t}\in\{0,1\}$, we define
\begin{equation}
    r_{u,t} = y_{u,t}\,(a_{u,t}-a_{\min}), \qquad a_{\min} := \min \mathcal{A}.
    \label{eq:reward}
\end{equation}
This reward is zero for failures and grows with item difficulty for successes, so the agent is incentivized to assign the hardest item the student can still solve. Under the Rasch model the conditional expected reward is
\begin{equation}
    \bar{r}(x,a) = (a-a_{\min})\,\sigma(x-a),
    \label{eq:expected_reward}
\end{equation}
Crucially, $r_{u,t}$ depends only on the observed outcome and the 
item's recorded difficulty, not on the student's proficiency, so 
it can be read off directly from the log without any additional modelling.

Although the realized reward $r_{u,t}$ is zero whenever a student fails, the
policy optimizes the \emph{expected} reward $\bar r(x,a)$ in Eq.~(2): $r_{u,t}$
is an unbiased single-sample estimate of $\bar r$, so an individual failure does
not imply a poorly chosen action. A difficulty that is well calibrated to a
student's proficiency has high expected reward regardless of any realized
outcome. Note also that $r_{u,t}$ is an internal optimization signal for
selecting difficulty, never a score shown to or imposed on the student. What
$\bar r$ captures is thus \emph{expected calibrated success}, i.e. placing students
at their challenge frontier, rather than a direct measure of learning gains.


\subsection{Adaptive behavior policy estimation}

The reduced-form behavior policy $\pi_t(a\mid z)$ captures the 
distribution of difficulty actually assigned to a student of true 
proficiency $z$ at time $t$. Crucially, this is \emph{not} the 
ITS's internal decision rule: the ITS does not observe $z_u$ 
directly, so it cannot condition on it explicitly. Rather, $\pi_t(a\mid z)$ summarizes the conditional
assignment pattern observed in the logs. This pattern may arise for two
reasons.
First, the ITS adapts based on observable proxies 
(e.g.\ recent performance) that correlate with $z_u$. Second, 
students themselves may exercise agency in choosing or attempting 
problems---and unlike the ITS, students may have direct access to 
their own knowledge. Both mechanisms produce a distribution of 
assigned difficulties that is conditional to $z$, even though 
neither agent observes it directly.

Estimating how $\pi_t(a\mid z)$ varies with $z$ and evolves with $t$ is
therefore a core contribution of this work. In light of the reward
defined above, a well-designed ITS should assign
difficulty in a way that is both adaptive and appropriately calibrated
to the student: more proficient students should receive harder items,
less proficient students easier ones, and the overall level should
shift upward as students progress through the sequence.\\ Thus,
$\pi_t(a\mid z)$ is informative not only about whether the system
adapts, but also about whether it tends to assign items that are too
easy, too difficult, or well matched to the student's level under the
desirable-difficulty objective. Conversely, a flat, diffuse, or poorly
aligned $\pi_t(a\mid z)$ suggests weak or miscalibrated adaptation.
The estimated behavior policy therefore provides an interpretable
diagnostic of how well the logged assignment rule aligns with the
reward objective, independently of any downstream policy learning.

Since $z_u$ is unobserved, we estimate 
this reduced-form policy by conditioning on the proxy 
$x = \hat{z}_u$, yielding $\pi_t(a \mid x)$. This introduces two 
error sources: a \emph{behavior-model error} (the parametric family 
may misfit the true conditional) and a \emph{knowledge estimation error} 
(we condition on $\hat{z}_u$ rather than $z_u$).

Once each interaction is represented as $(x_{u,t}, a_{u,t}, r_{u,t}, t)$, 
the tutoring log can be interpreted as a sample from a continuous stochastic contextual 
bandit problem and a time-dependent 
logging policy, making propensity estimation the central prerequisite for 
all offline estimators in Section~\ref{sec:ocbf}.

For each round $t$, we retrieve context--action pairs from all students with 
$T_u \ge t$,
\[
    \mathcal{D}^{(t)} := \{(x_{u,t},\,a_{u,t}) : T_u \ge t\},
\]
and approximate $\pi_t(a\mid x)$ by a parametric family 
$f^{(t)}_\eta(a\mid x)$, estimated by maximum likelihood on 
$\mathcal{D}^{(t)}$ alone:
\[
    \hat\eta_t \in \arg\max_\eta 
    \sum_{(x,a)\in \mathcal{D}^{(t)}} \log f^{(t)}_\eta(a \mid x).
\]

\begin{figure}[t]
    \centering
    \includegraphics[width=0.9\linewidth]{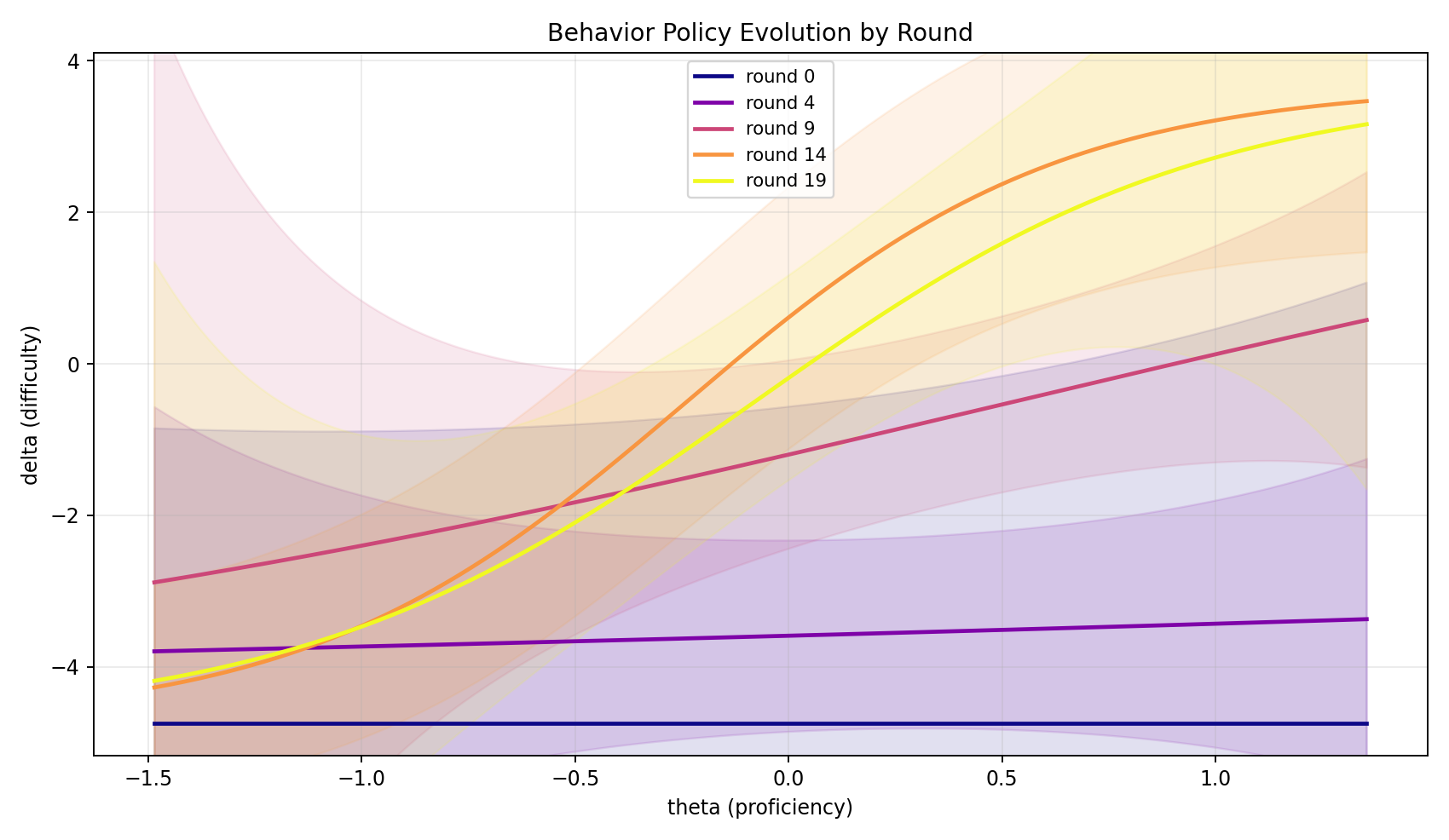}
    \caption{Estimated adaptive behavior policy \(\pi_t(a \mid x)\) at different steps of the learning trace. Shaded areas represent the variance.}
    \label{fig:adaptive_policy}
\end{figure}

The fitted propensities 
$\hat{p}_{u,t} := \hat{\pi}_t(a_{u,t}\mid x_{u,t})$ are then used in all 
off-policy estimators, while Figure~\ref{fig:adaptive_policy} illustrates 
the estimated policy at several rounds--directly visualizing how the 
collective assignment rule adapts to proficiency and shifts over time.

\section{Offline Contextual Bandit Framework}
\label{sec:ocbf}

\subsection{Logged dataset and context distribution shift}
In our setup, context $x_{u,t} = \hat{z}_u$ is a proficiency 
estimate derived from the student's complete interaction trace. Since 
round $t$ only includes students with $T_u \geq t$, later rounds 
systematically contain students with longer traces, which produces 
two compounding sources of context shift.

The first is a \emph{statistical estimation effect}: longer traces 
yield more precise Rasch estimates that are less attracted toward 
the center of the proficiency distribution. With few observations, 
regularization pulls $\hat{z}_u$ toward zero; with many 
observations, the estimate is free to reflect the student's true 
proficiency. As a result, $p_t(x)$ becomes progressively more 
dispersed at later rounds, with heavier tails and less mass near 
zero.

The second is a \emph{survivorship effect}: students who persist 
longer may not be a random subsample of the population. Depending 
on the system, more motivated or more proficient students may be 
over-represented in subsequent rounds, changing the mean of $p_t(x)$ 
and its spread.

\begin{figure}[t]
    \centering
    \includegraphics[width=0.9\linewidth]{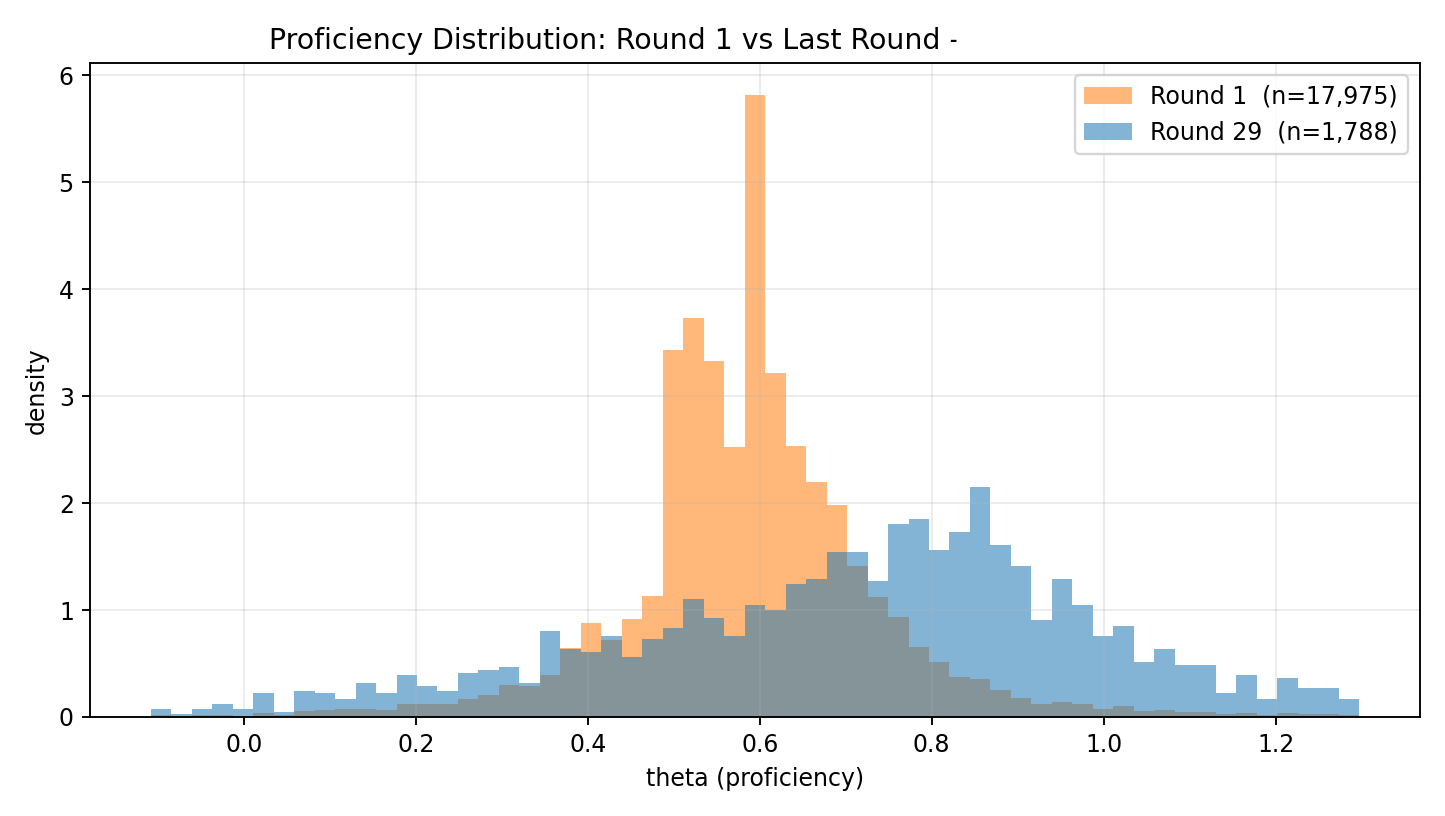}
    \caption{Evolution of the context distribution $p_t(x)$ across 
    rounds for \textsf{RoboMission}. At early rounds the distribution 
    is narrow and centered near zero; at later rounds it broadens 
    and shifts, reflecting both the improved precision of 
    longer-trace Rasch estimates and survivorship bias among 
    persisting students.}
    \label{fig:context_shift}
\end{figure}

Both effects cause $p_t(x)$ to drift substantially across rounds, 
as illustrated in Figure~\ref{fig:context_shift}. Let $p^\star(x)$ 
denote a reference context distribution. We augment each logged 
interaction with the reweighting factor
\[
    \omega_{u,t} := \frac{p^\star(x_{u,t})}{p_t(x_{u,t})},
\]
and the full logged dataset is
\[
    \mathcal{D}=\bigl\{
    (x_{u,t},\,a_{u,t},\,r_{u,t},\,\hat{p}_{u,t},\,
    \omega_{u,t},\,t)\bigr\}_{u,t}.
\]

\subsection{Policy value} We evaluate a stochastic target policy 
$\pi_\theta(a\mid x)$ whose value under the target context 
distribution is
\[
V^\star(\pi_\theta)
=
\int_{\mathcal X} p^\star(x)
\int_{\mathcal A} \pi_\theta(a\mid x)\,
\mathbb{E}[r\mid x,a]\,da\,dx.
\]
Since $\mathbb{E}[r\mid x,a]=(a-a_{\min})\,q(x,a)$, estimating 
this quantity directly requires a model for the correctness 
probability $q$. The two families of estimators below differ 
precisely in whether they commit to such a model.

\subsection{Model-free estimators} Defining the combined importance 
weight $w_{u,t}(\theta) := \omega_{u,t}\,{\pi_\theta(a_{u,t}\mid 
x_{u,t})}/{\hat{p}_{u,t}}$, the Importance Propensity Scoring (IPS), Self-Normalized-IPS (SNIPS), and Mixture Importance Sampling (MIS) estimators 
replace the unknown $\mathbb{E}[r\mid x,a]$ with the observed reward 
$r_{u,t}$:
\begin{align}
    \widehat{V}_{\mathrm{IPS}}  &= 
        \tfrac{1}{n}\textstyle\sum_{u,t} 
        w_{u,t}\,r_{u,t}, \label{eq:ips}\\
    \widehat{V}_{\mathrm{SNIPS}} &= 
        \textstyle\sum_{u,t} w_{u,t}\,r_{u,t} 
        \big/ \sum_{u,t} w_{u,t}, \label{eq:snips}\\
    \widehat{V}_{\mathrm{MIS}}   &= 
        \tfrac{1}{n}\textstyle\sum_{u,t} \omega_{u,t}\,
        \tfrac{\pi_\theta(a_{u,t}\mid x_{u,t})}
        {\bar\pi(a_{u,t}\mid x_{u,t})}\,r_{u,t}, 
        \label{eq:mis}
\end{align}
where $\bar\pi(a\mid x)=\sum_t\lambda_t\,\hat\pi_t(a\mid x)$ 
with $\lambda_t=n_t/n$ pools all rounds into a single denominator.\\ 
Crucially, none of these estimators requires an explicit model 
for $q$: they are agnostic to the functional form of the 
correctness probability and rely only on the observed binary 
outcomes encoded in $r_{u,t}$.

\subsection{Model-based and doubly robust estimators} 
A complementary strategy is to directly model 
$\mathbb{E}[r\mid x,a]$, which in our reward formulation amounts 
to specifying $q$. A natural choice is the Rasch model, giving
\[
    \hat{r}(x,a) := (a-a_{\min})\,\sigma(x-a).
\]
The direct method (DM) estimator then averages this reward model 
under the target policy:
\[
    \widehat{V}_{\mathrm{DM}}(\pi_\theta) = 
    \int_{\mathcal X} p^\star(x)
    \int_{\mathcal A}
    \pi_\theta(a\mid x)\,\hat{r}(x,a)\,da\,dx.
\]
Under this model the optimal difficulty level for a student of 
proficiency $x$ is
\[
    \delta^\star(x) 
    \in \arg\max_{a \in \mathcal{A}}\;
    (a - a_{\min})\,\sigma(x - a),
\]
which balances item difficulty against the probability of a 
correct response. The corresponding oracle policy is
\[
    \pi^\star_{\mathrm{IRT}}(a \mid x) 
    = \mathcal{N}\!\bigl(a;\,\delta^\star(x),\,\sigma^2_{\min}\bigr),
\]
where $\sigma^2_{\min}$ is a small fixed variance introduced to 
keep the policy stochastic---as required by importance 
sampling---while concentrating as much mass as possible around 
the optimal difficulty. It is the best policy achievable under the Rasch reward 
model, and any learned policy that approaches it under the DM 
estimator can be considered well-calibrated. DM has very low variance  but is necessarily  biased because the 
Rasch specification of $q$ is  always misspecified. The doubly robust 
estimator \cite{dudik2011doublyrobustpolicyevaluation} combines both approaches:
\begin{equation}
    \widehat{V}_{\mathrm{DR}} = \widehat{V}_{\mathrm{DM}} + 
    \tfrac{1}{n}\textstyle\sum_{u,t} 
    w_{u,t}\bigl(r_{u,t}-\hat{r}(x_{u,t},a_{u,t})\bigr),
    \label{eq:dr}
\end{equation}
and is consistent if either the behavior model or the reward 
model is correctly specified.

\subsection{Policy learning} Given any estimator $\widehat{V}$, 
the learned policy solves
\[
    \hat\theta \in \arg\max_{\theta\in\Theta}\;
    \widehat{V}(\pi_\theta),
\]
where $\pi_\theta$ is a Gaussian (or truncated Gaussian) density 
over the continuous difficulty space. Each estimator in Equations 
\eqref{eq:ips} to \eqref{eq:dr} induces a different learning 
objective, whose effects we compare in 
Section~\ref{sec:offline_policy_learning_results}.

\section{Experiments}
\label{sec:offline_policy_learning_results}

We evaluate offline contextual bandit learning on four datasets.
\textsf{RoboMission} is a dataset of block-based programming problems \cite{effenberger2018towards}.
\textsf{Assistments2015} and \textsf{Assistments2009} are two datasets
from the ASSISTments platform \cite{feng2009addressing}.
\textsf{Pix} is a dataset from an ITS that delivers problems linked to
digital competencies \cite{vie2017heuristic}. To avoid collapsing a long tail of interactions
into a single bucket, we truncate the horizon to 30 rounds for
\textsf{RoboMission} and \textsf{Pix}, and to 50 rounds for both
ASSISTments datasets.
Table~\ref{tab:datasets} summarizes the resulting dataset statistics.

\begin{table}[ht]
\centering
\caption{Dataset statistics. $[\delta_{\min}, \delta_{\max}]$
denotes the range of item difficulties on the latent scale
estimated by the Rasch model. \textsf{Assistments2009} is
restricted to the 8,000 most frequently attempted items.}
\small
\setlength{\tabcolsep}{4pt}
\begin{tabular}{lrrrrl}
\toprule
Dataset & Students & Interactions & Horizon &
    $[\delta_{\min},\, \delta_{\max}]$ \\
\midrule
RoboMission     & 20,556 & 258,670   & 30 & $[-3.700,\; 2.571]$ \\
Assistments2015 & 14,567 & 473,582   & 50 & $[-1.756,\; 0.709]$ \\
Assistments2009 &  4,066 &  97,571   & 50 & $[-1.292,\; 1.307]$ \\
Pix             & 100,000 & 2,978,909 & 30 & $[-4.751,\; 3.686]$ \\
\bottomrule
\end{tabular}
\label{tab:datasets}
\end{table}

\paragraph{Behavior and target policy class.}
Student proficiency is estimated by fitting a Rasch model on the full
interaction data, and we use the resulting latent ability estimate as
the context variable $x$. For each dataset and for each round, we fit
a behavior policy and learn a parametric target policy that recommends
items based on student proficiency. The target policy follows a Gaussian
distribution:
\[
    \pi_\theta(a \mid x) = \mathcal{N}\!\bigl(a;\;\mu_\theta(x),\,\sigma_\theta(x)^2\bigr),
\]
where the mean is linear in proficiency,
$\mu_\theta(x) = \beta_{\mu,0} + \beta_{\mu,1}\,x$,
and the log standard deviation is quadratic,
$\log \sigma_\theta(x)
= \beta_{\sigma,0} + \beta_{\sigma,1}\,x + \beta_{\sigma,2}\,x^2$,
leading to five learnable parameters
$\theta = (\beta_{\mu,0}, \beta_{\mu,1}, \beta_{\sigma,0},
\beta_{\sigma,1}, \beta_{\sigma,2})$. When the action space is
bounded to $[\delta_{\min}, \delta_{\max}]$, we use the corresponding
truncated Gaussian density and constrain $\mu_\theta(x)$ to stay in
range via a sigmoid reparameterization. We use the same family of distributions to fit the behavior policies $f_\eta$ in each round. 
We train one such policy per objective (IPS, SNIPS, DR, MIS), using
the target context distribution $p^\star$ estimated from the first
half of rounds. 

\paragraph{Evaluation protocol.}
Students are split 80/20 into train and test sets (split by student).
Round-specific behavior and target policies are fitted
exclusively on train students. Learned policies are then evaluated on
the held-out test set using the train-fitted behavior policy to estimate the
propensities used in the estimator: for each learned policy $\pi_\theta$, we compute
all five estimators (IPS, SNIPS, DR, MIS, DM) on the test set. The
behavior policy never sees test students during training: it is
learned from the train split and used to compute the propensities of context--actions tuples in the test split\footnote{Code is available at: \url{https://github.com/samgirr/counterfactual_learning_for_ITS}}.

\paragraph{ITS diagnostics and sanity checks.}
Before presenting policy-learning results, we report three
dataset-level diagnostics. Table~\ref{tab:behavior_fit} summarizes
 the behavior policies fit quality: higher RMSE indicates less accurate
propensity estimates and hence higher OPE variance.
Table~\ref{tab:diagnostics} reports two additional quantities: ITS
adaptivity via $\hat{V}(\pi_0)$ and $\hat{V}(\pi_{\text{last}})$, and
held-out Rasch calibration gap, where lower is better.

\begin{table}[t]
\centering
\caption{Behavior policy fit quality per dataset. RMSE measures the
deviation of the predicted mean difficulty $\hat\mu_t(x)$ from the
observed $\delta$. Higher RMSE indicates a harder-to-fit action
distribution, degrading propensity estimates and increasing OPE
variance.}
\small
\setlength{\tabcolsep}{4pt}
\begin{tabular}{lrrr}
\toprule
Dataset & RMSE (avg) & RMSE ($\hat\pi_{\text{last}}$) & Excluded \\
\midrule
RoboMission     & 0.931 & 1.025 & rounds 5--7 \\
Assistments2015 & 0.554 & 0.552 & none \\
Assistments2009 & 0.345 & 0.352 & none \\
Pix             & 0.704 & 0.681 & round 0 \\
\bottomrule
\end{tabular}
\label{tab:behavior_fit}
\end{table}

\begin{table}[t]
\centering
\caption{Dataset-level diagnostics used to contextualize the offline
policy learning results.}
\small
\begin{minipage}[t]{0.45\linewidth}
\centering
\setlength{\tabcolsep}{4pt}
\subcaption{ITS adaptivity diagnostic. Larger gaps indicate stronger
adaptation over the interaction horizon.}
\begin{tabular}{lrr}
\toprule
Dataset & $\hat{V}(\pi_0)$ & $\hat{V}(\pi_{\text{last}})$ \\
\midrule
Assistments2009 & 0.584 & 0.636 \\
Assistments2015 & 0.669 & 0.863 \\
RoboMission     & 1.619 & 1.877 \\
Pix             & 1.064 & 1.981 \\
\bottomrule
\end{tabular}
\label{tab:sanity_check}
\end{minipage}%
\quad
\begin{minipage}[t]{0.45\linewidth}
\centering
\subcaption{Calibration gap of the Rasch model on the test split. Lower is
better.\newline}
\setlength{\tabcolsep}{4pt}
\begin{tabular}{lr}
\toprule
Dataset & Calibration gap \\
\midrule
RoboMission     & 0.0212 \\
Assistments2015 & 0.0324 \\
Assistments2009 & 0.1112 \\
Pix             & 0.0280 \\
\bottomrule
\end{tabular}
\label{tab:rasch_calibration_gap}
\end{minipage}

\label{tab:diagnostics}
\end{table}

Behavior-model fit varies substantially across datasets, from $0.345$
on \textsf{Assistments2009} and $0.554$ on
\textsf{Assistments2015} to $0.704$ on \textsf{Pix} and $0.931$ on
\textsf{RoboMission}. For \textsf{RoboMission}, rounds~5--7 were
excluded because of degenerate fits (RMSE $> 19$), likely due to a
bimodal difficulty distribution with a spike at $\delta_{\min}$. For
\textsf{Pix}, round~0 was excluded (RMSE $=12.34$) because the ITS
starts with a deterministic assignment rule that concentrates all
students on the same item.

Taken together, these diagnostics provide context for the results that
follow. \textsf{Assistments2015} combines good fit with clear
adaptation, \textsf{Assistments2009} shows good behavior-model fit but
limited adaptation and substantially worse Rasch calibration, and
\textsf{RoboMission} and \textsf{Pix} exhibit stronger adaptation in a
more challenging propensity-estimation setting.

\begin{figure*}[t]
\centering
\begin{subfigure}[t]{0.49\textwidth}
    \centering
    \includegraphics[width=\textwidth]{%
        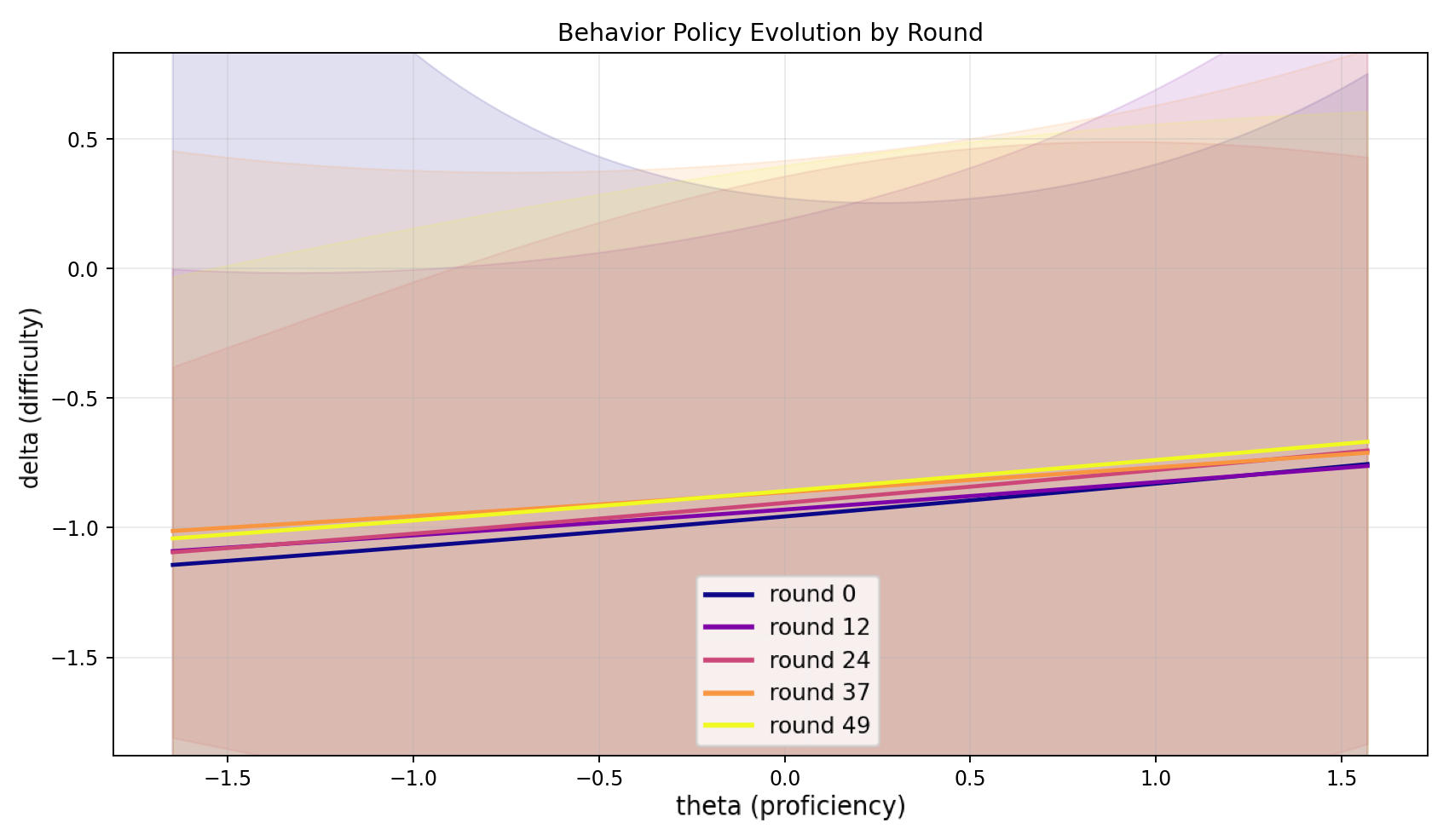}
    \caption{Behavior policy evolution.}
\end{subfigure}
\hfill
\begin{subfigure}[t]{0.49\textwidth}
    \centering
    \includegraphics[width=\textwidth]{%
        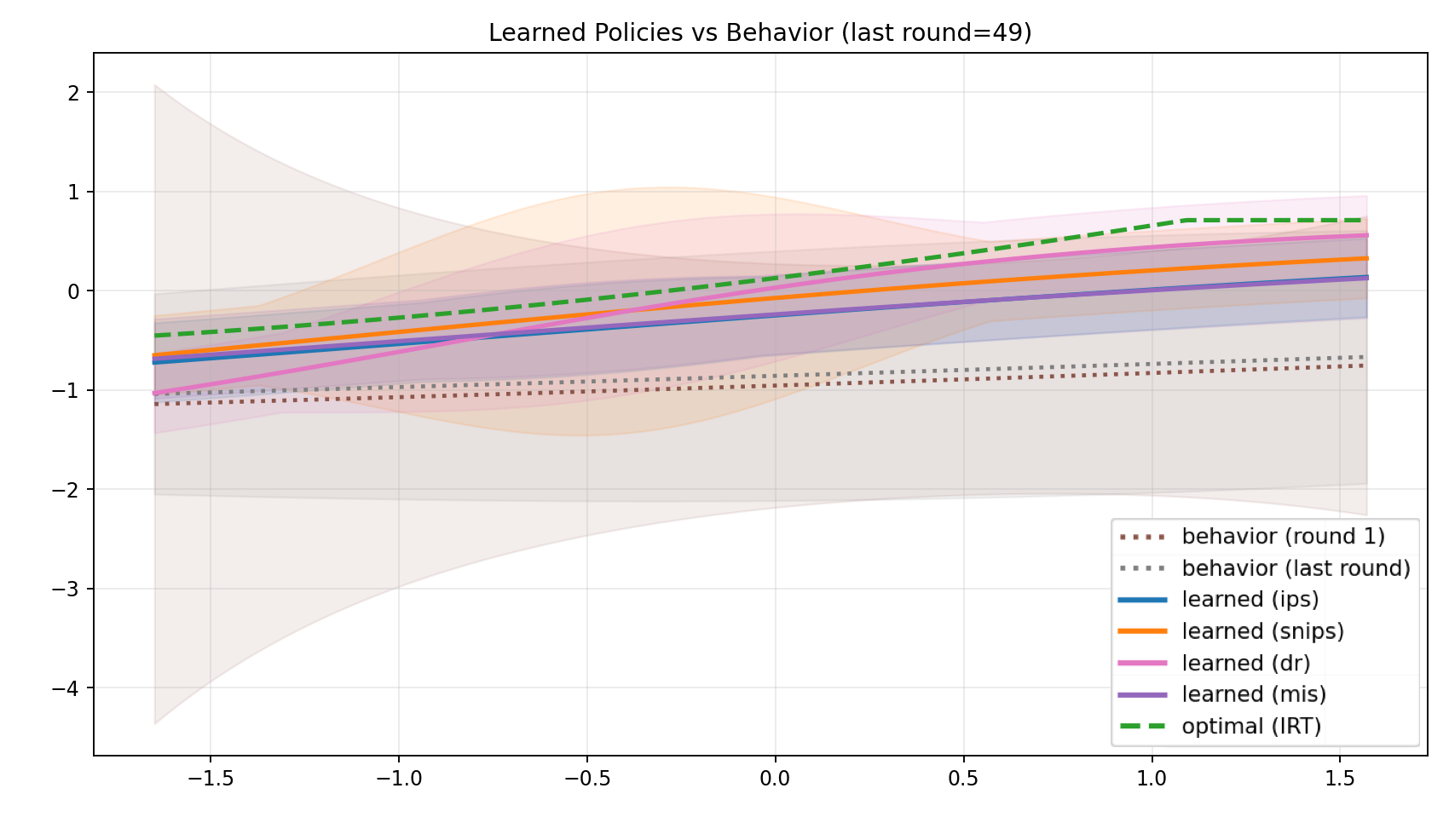}
    \caption{Learned policy curves.}
\end{subfigure}
\caption{\textsf{Assistments2015}. Left: consistent upward shift
across rounds without strong drift, producing good overlap across
the full interaction horizon. Right: all learned policies improve
substantially over the behavior baseline, with strong agreement
across estimators.}
\label{fig:skillbuilder}
\end{figure*}

\paragraph{\textsf{Assistments2015}.}
This dataset represents the most favorable setting for offline
contextual bandit learning: the behavior policy is adaptive 
horizon (see Table~\ref{tab:sanity_check}), yet drift is moderate enough
that overlap remains strong throughout. The behavior policy curves in
Figure~\ref{fig:skillbuilder}(a) show a consistent upward shift
across rounds without abrupt changes.

All learned policies improve substantially over the behavior baseline
(Table~\ref{tab:results_skillbuilder}). Among learned policies,
Learned~DR achieves the best SNIPS ($1.093$, a $23.8\%$ improvement
over the behavior policy), while the IRT oracle dominates the DR and
DM columns as expected. The consistency across estimators gives high
confidence that these gains are genuine.

\begin{table}[t]
\centering
\caption{Held-out test results on \textsf{Assistments2015}.}
\small
\setlength{\tabcolsep}{4pt}
\begin{tabular}{llrrrrr}
\toprule
Policy & Obj. & IPS & SNIPS & DR & MIS & DM \\
\midrule
Behavior      & --    & 0.881 & 0.882 & 0.877 & 0.883 & 0.883 \\
IRT oracle    & --    & 0.804 & 1.083 & \textbf{1.114} & 0.821 & \textbf{1.170} \\
Learned IPS   & ips   & \textbf{1.270} & 1.080 & 1.076 & \textbf{1.268} & 1.080 \\
Learned SNIPS & snips & 1.231 & 1.089 & 1.088 & 1.230 & 1.096 \\
Learned DR    & dr    & 1.175 & \textbf{1.093} & 1.099 & 1.176 & 1.119 \\
Learned MIS   & mis   & \textbf{1.270} & 1.080 & 1.076 & 1.268 & 1.081 \\
\bottomrule
\end{tabular}
\label{tab:results_skillbuilder}
\end{table}

\begin{figure*}[t]
\centering
\begin{subfigure}[t]{0.49\textwidth}
    \centering
    \includegraphics[width=\textwidth]{%
        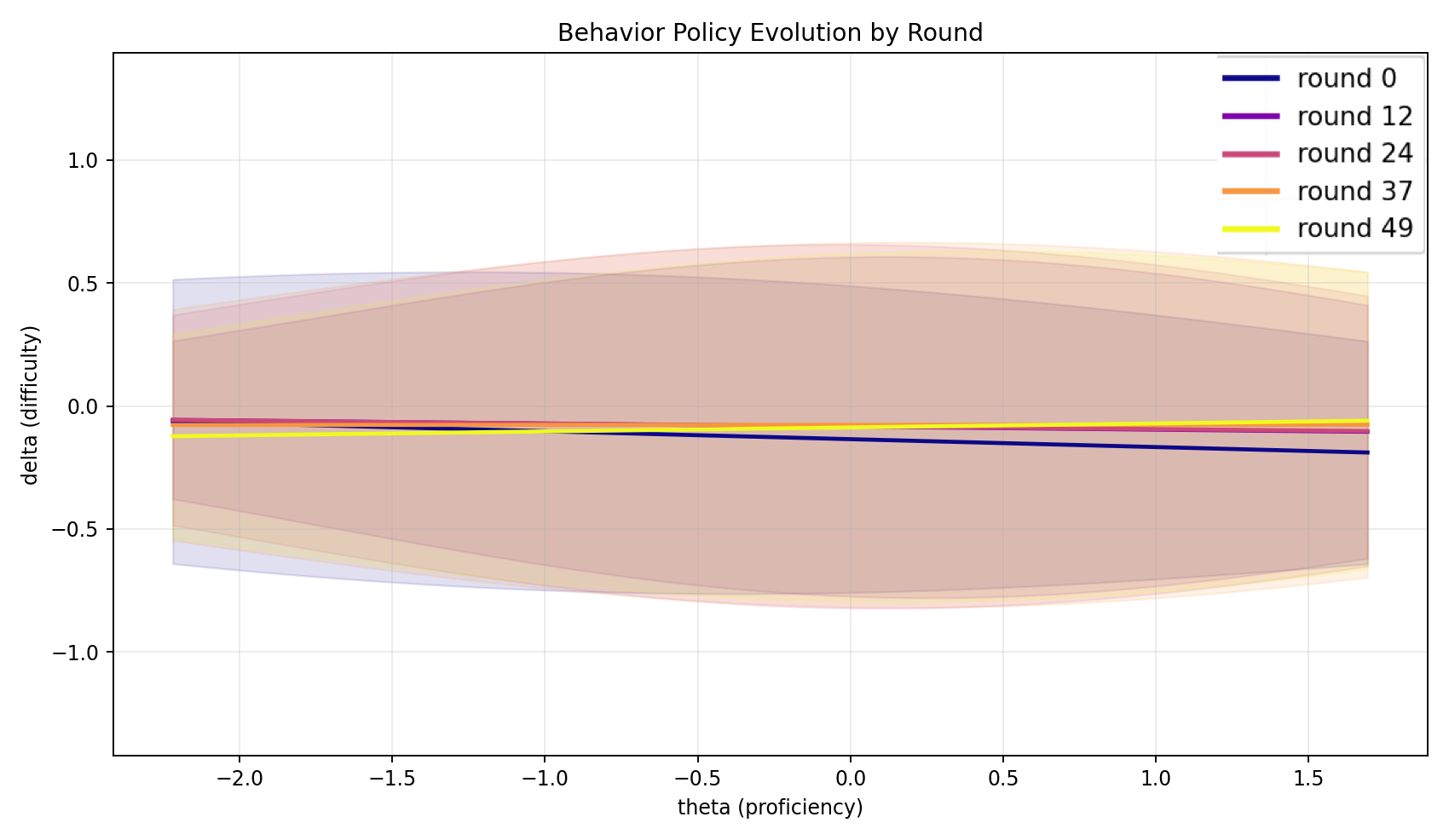}
    \caption{Behavior policy evolution.}
\end{subfigure}
\hfill
\begin{subfigure}[t]{0.49\textwidth}
    \centering
    \includegraphics[width=\textwidth]{%
        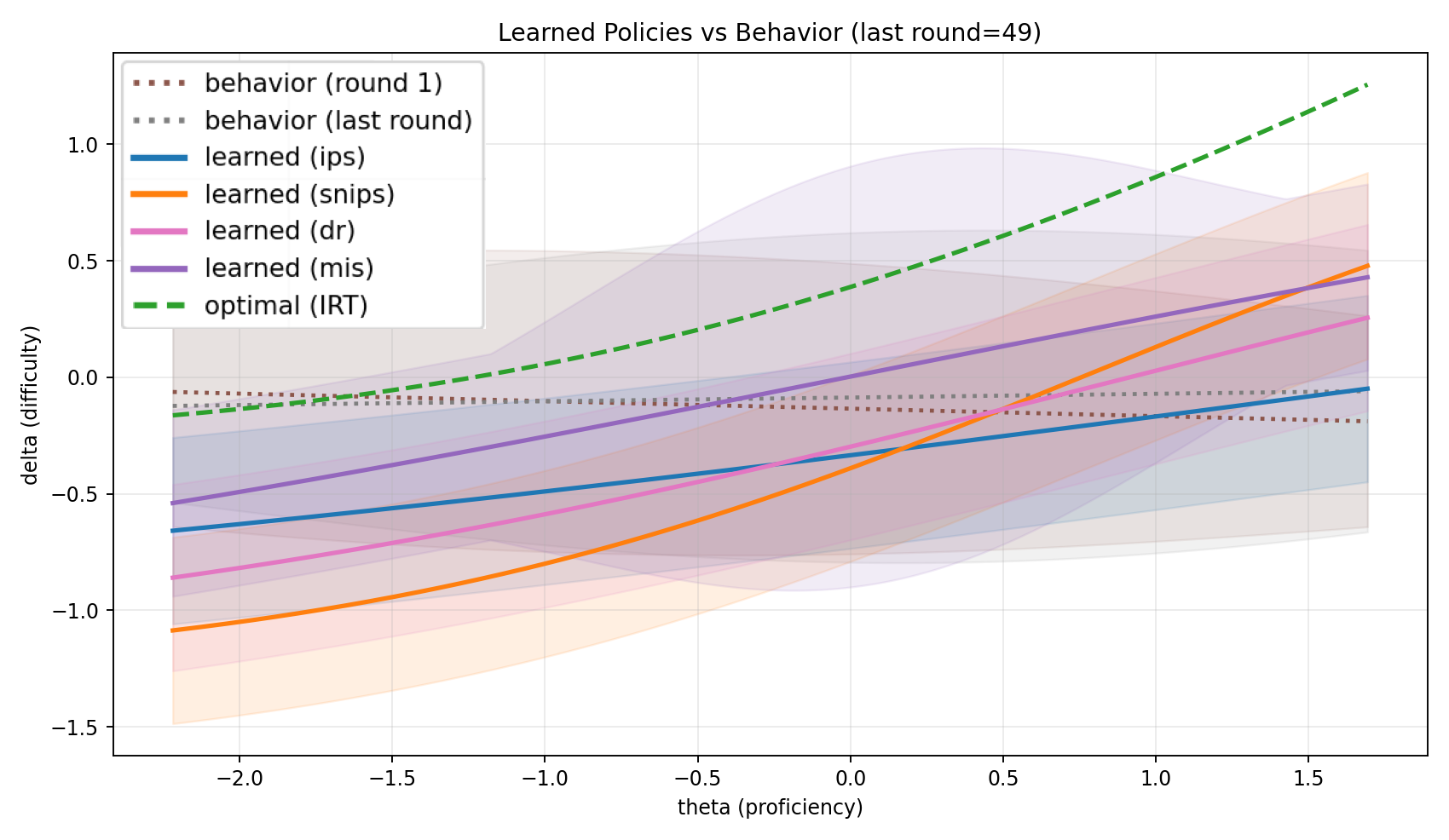}
    \caption{Learned policy curves.}
\end{subfigure}
\caption{\textsf{Assistments2009}. Left: near-flat behavior policy
curves across rounds and proficiency levels, indicating poor
adaptivity. Right: learned policies remain close to the behavior
baseline, reflecting the limited support available for policy
improvement.}
\label{fig:assistments2009}
\end{figure*}

\paragraph{\textsf{Assistments2009}.}
The behavior policy shows only modest adaptiveness: the con\-text-weighted
reward improves from $0.584$ to $0.636$ over the full horizon
(Table~\ref{tab:sanity_check}), and the behavior policy curves in
Figure~\ref{fig:assistments2009}(a) are nearly flat across all rounds
and proficiency levels. Overlap is good, but the difficulty lies in
the limited exploration: the system assigns a narrow range of
difficulties regardless of student ability, leaving little support for
learning a substantially different policy. An additional issue on this
dataset is Rasch miscalibration. Table~\ref{tab:rasch_calibration_gap}
shows that \textsf{Assistments2009} has the largest calibration
gap ($0.1112$) among all datasets. This helps explain the mismatch
between the IRT oracle's SNIPS estimate ($0.537$), which is below the
behavior policy ($0.641$), and its DM estimate ($0.787$), which is
much higher: the Rasch model appears to predict gains from more
difficult assignments that are not supported by the logged data.

Learned~SNIPS and Learned~DR achieve modest gains over the behavior
baseline (Table~\ref{tab:results_assistments2009}), while Learned~IPS
and Learned~MIS fail to improve reliably. The learned policy curves
confirm this pattern: all objectives remain close to the behavior
baseline, with limited room to diverge
(Figure~\ref{fig:assistments2009}(b)).

\begin{table}[t]
\centering
\caption{Held-out test results on \textsf{Assistments2009}.}
\small
\setlength{\tabcolsep}{4pt}
\begin{tabular}{llrrrrr}
\toprule
Policy & Obj. & IPS & SNIPS & DR & MIS & DM \\
\midrule
Behavior      & --    & 0.653 & 0.641 & 0.641 & 0.648 & 0.651 \\
IRT oracle    & --    & 0.440 & 0.537 & 0.623 & 0.444 & \textbf{0.787} \\
Learned IPS   & ips   & 0.652 & 0.627 & 0.623 & 0.655 & 0.696 \\
Learned SNIPS & snips & \textbf{0.696} & \textbf{0.686} & \textbf{0.693} & \textbf{0.694} & 0.658 \\
Learned DR    & dr    & 0.692 & 0.684 & 0.692 & 0.690 & 0.665 \\
Learned MIS   & mis   & 0.646 & 0.620 & 0.615 & 0.652 & 0.709 \\
\bottomrule
\end{tabular}
\label{tab:results_assistments2009}
\end{table}

\begin{figure*}[t]
\centering
\begin{subfigure}[t]{0.49\textwidth}
    \centering
    \includegraphics[width=\textwidth]{%
        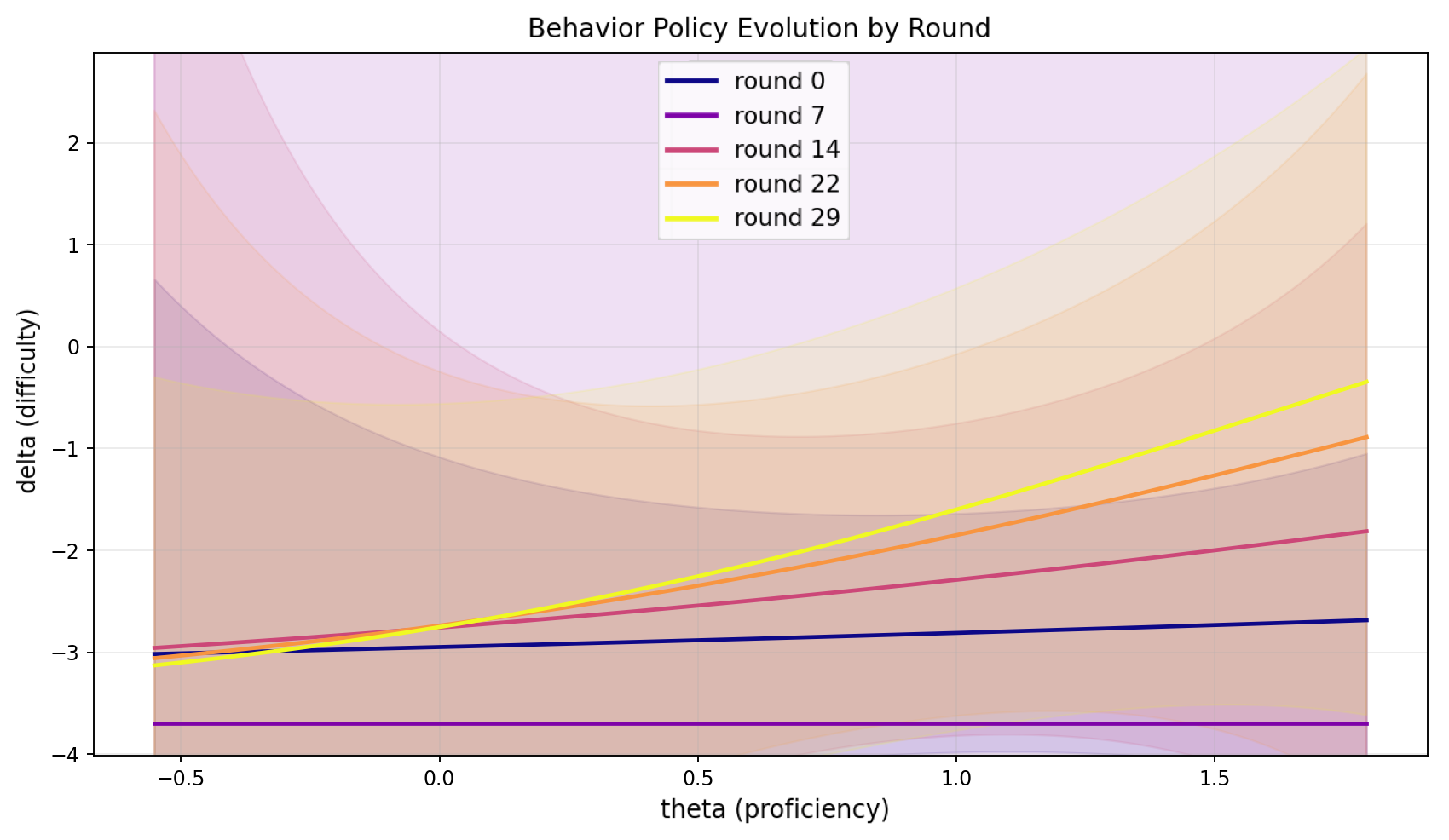}
    \caption{Behavior policy evolution.}
\end{subfigure}
\hfill
\begin{subfigure}[t]{0.49\textwidth}
    \centering
    \includegraphics[width=\textwidth]{%
        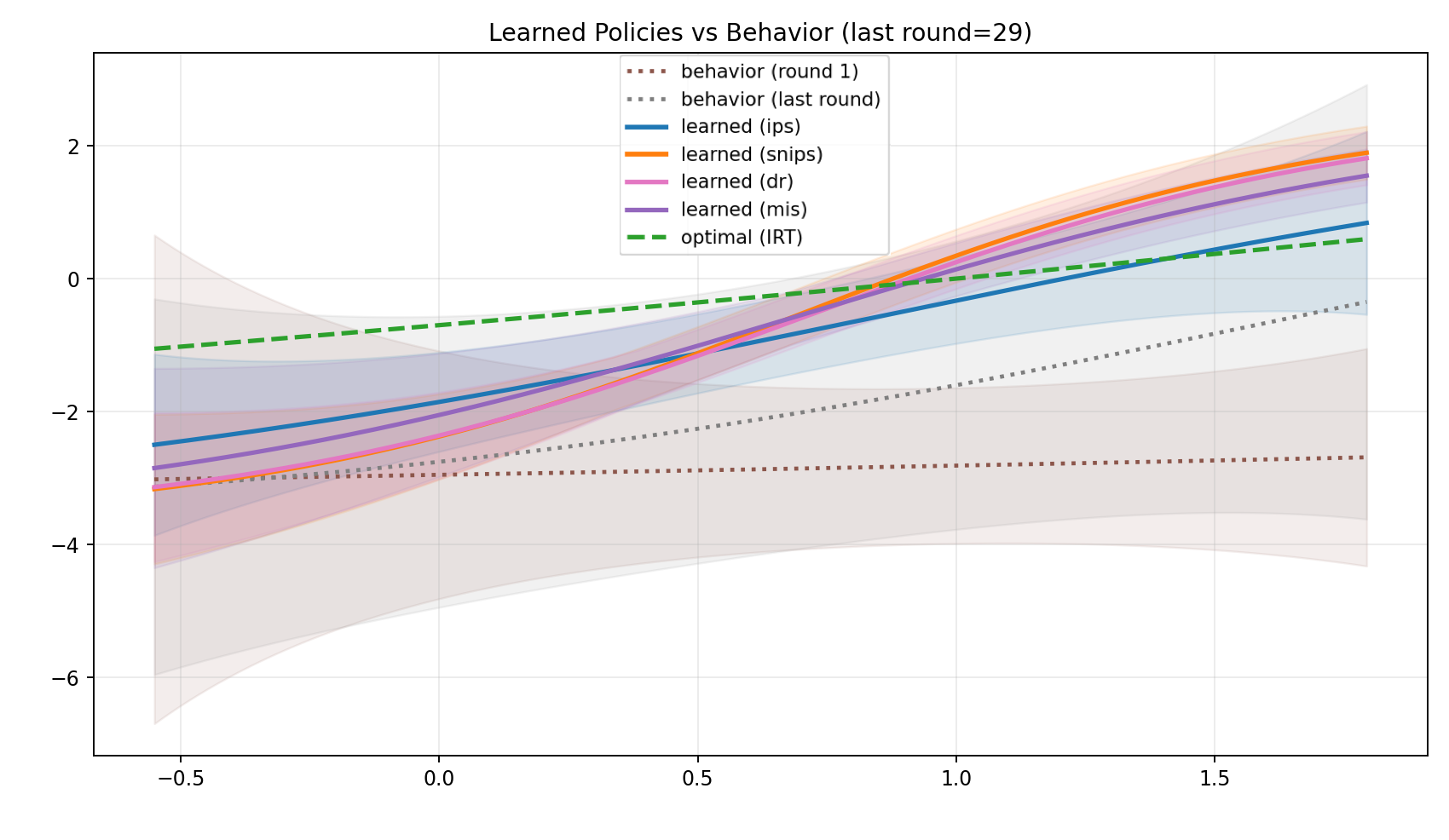}
    \caption{Learned policy curves.}
\end{subfigure}
\caption{\textsf{RoboMission}. Left: strong upward shift across
rounds indicating high ITS adaptiveness, but also large
round-to-round drift that degrades behavior model fit at certain
rounds. Right: all learned policies improve over the behavior
baseline, with SNIPS and DR being the most reliable metrics.}
\label{fig:robomission}
\end{figure*}

\paragraph{\textsf{RoboMission}.}
The ITS shows strong adaptation over the
horizon: $\hat{V}(\pi_0)$ rises from $1.619$ to
$\hat{V}(\pi_{\text{last}})=1.877$
(Table~\ref{tab:sanity_check}), and the behavior policy curves in
Figure~\ref{fig:robomission}(a) shift upward across rounds, assigning
harder items to more proficient students. At the same time,
\textsf{RoboMission} has the highest average RMSE ($0.931$;
Table~\ref{tab:behavior_fit}), indicating that the behavior policy is
 harder to model accurately than in the other datasets.
This poor fit helps explain the discrepancy between the reference value
$\hat{V}(\pi_{\text{last}})$ in Table~\ref{tab:sanity_check} and the
behavior-policy estimates in
Table~\ref{tab:results_attempts}.

All learned policies improve over the behavior baseline under SNIPS and
DR (Table~\ref{tab:results_attempts}), with Learned~MIS achieving the
best results, a $22.5\%$ improvement in SNIPS over the behavior policy.

\begin{table}[t]
\centering
\caption{Held-out test results on \textsf{RoboMission}.}
\small
\setlength{\tabcolsep}{4pt}
\begin{tabular}{llrrrrr}
\toprule
Policy & Obj. & IPS & SNIPS & DR & MIS & DM \\
\midrule
Behavior      & --    & 1.968 & 1.981 & 2.006 & 2.139 & 2.013 \\
IRT oracle    & --    & 1.694 & 2.370 & 2.408 & 2.563 & \textbf{2.443} \\
Learned IPS   & ips   & 2.332 & 2.360 & 2.361 & 2.749 & 2.241 \\
Learned SNIPS & snips & 2.347 & 2.409 & 2.408 & 2.858 & 2.274 \\
Learned DR    & dr    & 2.315 & 2.425 & 2.431 & 2.875 & 2.305 \\
Learned MIS   & mis   & 2.302 & \textbf{2.426} & \textbf{2.432}
    & \textbf{2.868} & 2.309 \\
\bottomrule
\end{tabular}
\label{tab:results_attempts}
\end{table}



\paragraph{\textsf{Pix}.}
The behavior policy is the most strongly adaptive across all datasets:
the context-weighted reward nearly doubles from $1.064$ to $1.981$
(Table~\ref{tab:sanity_check}), the largest absolute gain observed.
This strong drift is reflected in the behavior policy evolution shown in
Figure~\ref{fig:adaptive_policy}. The behavior model RMSE is
intermediate (Table~\ref{tab:behavior_fit}), with all rounds after
round~0 well fitted.

All learned policies improve substantially over the behavior baseline
(Table~\ref{tab:results_pix}), with Learned~DR achieving the best
SNIPS and DR among all policies --- a $38.9\%$ improvement in SNIPS,
the largest gain across all datasets. The IRT oracle again dominates
the DM column.

Comparing the behavior-policy IPS and SNIPS estimates in
Table~\ref{tab:results_pix} with the reference value
$\hat{V}(\pi_{\text{last}})=1.981$ from
Table~\ref{tab:sanity_check} reveals an instructive asymmetry on this
dataset: IPS underestimates the final-round policy value, whereas
SNIPS overestimates it.
This is a
direct consequence of the strong behavior drift. The importance weights
for evaluating the last-round policy $\hat\pi_{29}$ on data from an
early round~$t$ are proportional to
$\hat\pi_{29}(a\mid x)/\hat\pi_t(a\mid x)$. At early rounds, the
system assigned much easier items than at round~29, so
$\hat\pi_{29}$ assigns them low probability and the weights are small.
IPS divides by the total number of samples, so these low-weight
early-round observations still occupy a large fraction of the
denominator, pulling the estimate down. SNIPS normalises by the sum
of weights instead, effectively concentrating mass on late rounds
where weights are close to~1 and rewards are high, which inflates the
estimate. The more adaptive the behavior policy, the larger this
divergence between IPS and SNIPS --- a pattern that is mild on the
ASSISTments datasets and most pronounced here.

\begin{table}[t]
\centering
\caption{Held-out test results on \textsf{Pix}.}
\small
\setlength{\tabcolsep}{4pt}
\begin{tabular}{llrrrrr}
\toprule
Policy & Obj. & IPS & SNIPS & DR & MIS & DM \\
\midrule
Behavior      & --    & 1.781 & 2.171 & 2.136 & 2.132 & 2.151 \\
IRT oracle    & --    & 2.227 & 2.860 & 2.886 & 2.642 & \textbf{2.734} \\
Learned IPS   & ips   & 2.793 & 2.995 & 2.979 & 3.085 & 2.657 \\
Learned SNIPS & snips & 2.678 & 3.012 & 2.990 & 2.956 & 2.690 \\
Learned DR    & dr    & 2.731 & \textbf{3.016} & \textbf{2.999}
    & 3.006 & 2.695 \\
Learned MIS   & mis   & 2.678 & 2.907 & 2.887 & \textbf{3.125} & 2.629 \\
\bottomrule
\end{tabular}
\label{tab:results_pix}
\end{table}




\section{Conclusion}

We presented an offline contextual bandit framework for learning
instructional policies directly from logged educational data,
without interaction with students or reliance on a student
simulator. Experiments on four real-world datasets show that learned
policies consistently improve over the behavior baseline out of
sample, with SNIPS and DR emerging as the most reliable objectives
across datasets and estimators. Our behavior-policy diagnostic also
reveals meaningful differences in ITS adaptiveness across systems.
More broadly, we show that Rasch model calibration directly affects
the reliability of offline evaluation, and should therefore be
reported alongside any such analysis.

\paragraph{Deployment and cold start.}
Our evaluation conditions on $x_{u,t}=\hat z_u$ estimated from each student's
\emph{complete} trace, which suits offline analysis but is unavailable when
serving a new student. At deployment, a student with no history is initialized
at the population average ($\hat z\approx 0$); as they answer items, $\hat z_u$
is re-estimated online from the growing prefix and the policy is replayed on the
updated context. The learned policy is thus only as good as this running
estimate, which is least precise in the earliest rounds---where regularization
pulls $\hat z_u$ toward zero and sharpens as evidence
accumulates. This online protocol bridges our full-trace evaluation context and
a realizable deployment loop, at the cost of degraded estimates during a
student's first interactions. 

\paragraph{Limitations and future work.}
Our bandit formulation optimizes immediate rewards rather than
long-term learning gains. Extending the framework to sequential
decision-making would allow richer policies, but at substantially
greater data and computational cost~\cite{doroudi2017robust}. In
addition, our behavior-policy estimation currently relies on simple
parametric families. A promising direction is to consider more
flexible models, such as mixture distributions, which could better
capture complex assignment patterns in the logs. This is especially
relevant at later rounds, where the behavior policy may become
multimodal, for example if the ITS separates students
into distinct difficulty tracks. More expressive behavior models could
therefore improve both interpretability and the stability of
downstream off-policy evaluation. When the target policy departs far
from the behavior policy, estimators disagree and effective sample
size drops; we advocate for logging policies that include deliberate
exploration and for releasing behavior-policy descriptions alongside
datasets~\cite{saito2021open}. Finally, visualizing both behavior and
learned policies as functions of proficiency opens the door to
fairness analyses~\cite{gardner2019evaluating} and may also help
identify new knowledge-tracing models from the structure of the
learned policies themselves.

\begin{credits}
\subsubsection{\ackname} This study was funded by Île-de-France, Paris Region PhD.

\subsubsection{\discintname}
Samuel Girard has received a research grant from public group interest Pix.

\subsubsection{Use of Generative AI.} 
Generative AI tools were used for coding assistance, language refinement, and grammatical editing. All intellectual content, including research design, analysis, and interpretation, was performed by the authors.

\subsubsection{Ethics and Societal Aspects.} The research conducted in this paper is solely made on offline pseudonymized data in order to benefit human learning. The purpose of off-policy learning is precisely to learn promising strategies to ask questions to humans while avoiding to rely on actual online experimentations on humans.

\end{credits}

\bibliographystyle{splncs04}
\bibliography{sample-base}

\end{document}